%% file: main.tex
\documentclass{article}

\PassOptionsToPackage{numbers, compress}{natbib}
\usepackage[preprint]{neurips_2025}

\workshoptitle{MATH-AI}

\usepackage[utf8]{inputenc} 
\usepackage[T1]{fontenc}    
\usepackage{hyperref}       
\usepackage{float}
\usepackage{flafter}
\usepackage{url}            
\usepackage{multirow}
\usepackage{booktabs}   

\usepackage{placeins}
\usepackage{amsfonts}       
\usepackage{amsmath}
\usepackage{subcaption}
\usepackage{graphicx}
\usepackage{nicefrac}       
\usepackage{microtype}      
\usepackage{xcolor}         
\usepackage{enumitem}

\definecolor{cadmiumgreen}{rgb}{0.0, 0.42, 0.24}
\definecolor{cornellred}{rgb}{0.7, 0.11, 0.11}
\hypersetup{
  linkcolor = cornellred,
  citecolor  = cadmiumgreen,
  colorlinks = true,
}

\title{Revisiting the Uniform Information Density Hypothesis in LLM Reasoning Traces}

\author{%
Minju Gwak \\
Department of Artificial Intelligence \\
Yonsei University \\
\texttt{mjgwak@yonsei.ac.kr} 
\And
Guijin Son \\
OneLine AI \\
\texttt{guijin.son@oneline.com}
\And
Jaehyung Kim \\
Department of Artificial Intelligence \\
Yonsei University \\
\texttt{jaehyungk@yonsei.ac.kr} 
}
\usepackage{multirow}
\begin{document}

\maketitle

\begin{abstract}
Large language models (LLMs) often solve problems using step-by-step Chain-of-Thought (CoT) reasoning, yet these intermediate steps are frequently unfaithful or hard to interpret. Inspired by the Uniform Information Density (UID) hypothesis in psycholinguistics -- which posits that humans communicate by maintaining a stable flow of information -- we introduce entropy-based metrics to analyze the information flow within reasoning traces. Surprisingly, across three challenging mathematical benchmarks, we find that successful reasoning in LLMs is globally non-uniform: correct solutions are characterized by uneven swings in information density, in stark contrast to human communication patterns. This result challenges assumptions about machine reasoning and suggests new directions for designing interpretable and adaptive reasoning models.
\end{abstract}

\input{1_intro}
\input{2_related_work}
\input{3_method}
\input{5_results}
\input{6_conclusion}

\bibliographystyle{unsrtnat}
\bibliography{neurips_2025}

\input{7_appendix}

\end{document}

%% file: 1_intro.tex
\section{Introduction}

Chain-of-Thought (CoT) reasoning has emerged as a central technique for improving large language models (LLMs) on complex reasoning tasks \citep{wei2023chainofthoughtpromptingelicitsreasoning, kojima2023largelanguagemodelszeroshot, chae2023dialoguechainofthoughtdistillationcommonsenseaware}. 
By generating step-by-step rationales, CoT allows models to decompose problems and produce more interpretable outputs \citep{golovneva2023roscoesuitemetricsscoring, prasad2023recevalevaluatingreasoningchains}. 
However, recent studies have highlighted the fragility of this approach \citep{zhao2025chainofthoughtreasoningllmsmirage}.
Specifically, despite generating longer reasoning traces, LLMs often fail to generalize, and their intermediate steps can be logically inconsistent or incoherent \citep{shojaee2025illusionthinkingunderstandingstrengths}. 
This raises an important question: \textit{how can we tell when LLMs are reasoning effectively, rather than merely generating superficially coherent text?}

Human communication provides a potential clue. A psycholinguistic theory suggests that effective communication relies on a uniform flow of information\citep{meister2021revisitinguniforminformationdensity, tsipidi2024surpriseuniforminformationdensity}, where ideas are expressed at a stable rate to match human cognitive processing limits. When information is delivered too unevenly, understanding breaks down. We hypothesize that a similar principle applies to LLM reasoning: just as humans produce language with balanced information flow, effective reasoning traces may exhibit comparable uniformity. To explore this link, we draw on cognitive science and psycholinguistics; for instance, \citet{bhambri2025cognitivelyinterpretablereasoningtraces} shows that reasoning paths interpretable to humans are also easier for models to generate and learn, suggesting a shared structure between human cognition and machine reasoning. To illustrate, a well-reasoned math solution might show consistent step-level progress, where each step builds smoothly on the previous one, while an incoherent solution might jump between overly trivial and overly complex steps.

We analyze the information flow of LLM-generated reasoning traces on challenging mathematical benchmarks. 
We define per-step measurements of information density, and examine by answer correctness. 
Then, we introduce three complementary metrics that quantify the uniformity entire reasoning trace, using entropy-based per-step measurement. 
Our experiments reveal a clear pattern: unlike human communication, reasoning traces with low global uniformity tend to produce correct answers. 
This suggests that effective reasoning balances local uniformity and low global uniformity.

Overall, our contributions are threefold:
\begin{itemize}[leftmargin=5.5mm,topsep=0pt]
    \item[$\circ$] To our knowledge, we are the first to introduce information-theoretic metrics for quantifying reasoning structure at both the step and trace level.
    \item[$\circ$] We find that reasoning patterns characterized by low global uniformity, correlate with reasoning success on challenging mathematical reasoning benchmarks.
    \item[$\circ$] We show that deviations from such patterns can serve as an internal signal for predicting failure cases, enabling potential improvements in LLM reasoning and evaluation.
\end{itemize}

%% file: 2_related_work.tex
\section{Related Work}

\subsection{Fragility of CoT and the role of individual reasoning steps}
CoT prompting improves reasoning but remains fragile ~\citep{wei2023chainofthoughtpromptingelicitsreasoning, zhao2025chainofthoughtreasoningllmsmirage}. 
Small, seemingly irrelevant perturbations in the reasoning chain can sharply reduce accuracy \citep{mirzadeh2025gsmsymbolicunderstandinglimitationsmathematical,tang2023largelanguagemodelsincontext}, suggesting that models often produce the appearance of reasoning rather than logically sound traces \citep{shojaee2025illusionthinkingunderstandingstrengths}. 
Moreover, longer reasoning steps do not necessarily reflect the true difficulty of the problem, and many intermediate steps can be altered or even removed without changing the final answer \citep{lanham2023measuringfaithfulnesschainofthoughtreasoning}. 
This raises doubts about the necessity and faithfulness of these step-by-step explanations.
Another line of recent research takes a different perspective: rather than viewing all steps as equally important, it suggests that a small subset of pivotal steps within CoT traces disproportionately drives predictions \citep{bogdan2025thoughtanchorsllmreasoning}.
Attribution methods and their frameworks identify and highlight these critical steps, emphasizing the need to understand how individual steps shape outcomes\citep{golovneva2023roscoesuitemetricsscoring, wu2023analyzingchainofthoughtpromptinglarge, bigelow2024forkingpathsneuraltext}. 
Despite these advances, there remains no clear interpretation of what constitutes a truly good reasoning pattern.

\subsection{Intrinsic signals in LLM reasoning}
Research on LLM reasoning has increasingly turned to internal model signals to gain insights into how reasoning unfolds.
Many approaches use these signals to improve performance, such as using self-consistency \citep{zuo2025ttrltesttimereinforcementlearning}, self-certainty \citep{kang2025scalablebestofnselectionlarge, zhao2025learningreasonexternalrewards}, or confidence to refine outputs, or using entropy-based measures to encourage diverse reasoning paths \citep{zhang2025rightquestionhalfanswer, agarwal2025unreasonableeffectivenessentropyminimization, gao2025oneshotentropyminimization, lee2025trainingfreellmverificationrecycling}. We shift focus from controlling reasoning with internal signals to understanding it through their structure. We ground our analysis in long-standing psycholinguistic theory to understand the structure of reasoning itself by revealing how information is introduced, transformed, and propagated through the reasoning process. 
Our step-level focus provides a deeper understanding of what constitutes a coherent reasoning trace, going beyond prior approaches that emphasize performance gains over interpretability.

%% file: 3_method.tex
\section{Exploring the UID Hypothesis in Reasoning Models}
\subsection{Background: Uniform information density hypothesis}
The Uniform Information Density (UID) hypothesis models language as a signal transmitted through a noisy channel with limited capacity \citep{meister2021revisitinguniforminformationdensity, tsipidi2024surpriseuniforminformationdensity}. 
It posits that speakers aim to convey information efficiently without overwhelming the listener's processing resources. 
Let an utterance $\textbf{u}$ = $[u_1$,$u_2$, \dots, $u_N]$ be a sequence of $N$ linguistic units, such as words, subwords, or characters, depending on the granularity of representation. 
For each unit $u_n$, we can define surprise as the unexpectedness of a unit, given its previous context. 
Formally, surprisal is defined as:
\begin{equation*}
    s(u_n) = -\log P(u_n \mid \textbf{u}_{<n}),
\end{equation*}
where $P(u_n |u_{<n})$ is the probability of seeing unit utterance $u_n$ after the earlier sequence $\textbf{u}_{<n}=[u_1,\dots,u_{n-1}]$. 
High surprisal of the unit denotes that it is very unexpected and hard to process for the person receiving the information, while units with lower surprisal are easier to process. 
In this sense, surprisal can be perceived as information content. 
To capture the overall cognitive load of a message, we aggregate this surprisal across all units in the sequence. 
Given a sequence of utterance $\textbf{u}$, the total processing effort can be expressed as:
\begin{equation*}
    \mathrm{Processing Effort}(u) \propto \sum_{n=1}^{N} s(u_n).
\end{equation*}
If information is concentrated in a few highly surprising units, the receiver experiences sharp spikes in processing difficulty; if it is too sparse, communication becomes inefficient. 
The high-level intuition of the UID hypothesis is that the most efficient strategy is to distribute surprisal as evenly as possible across the sequence, maintaining a stable level of processing effort. 
This tendency has been empirically observed across syllables, words, syntax, and discourse.

While UID has been extensively validated in human language, its implications for machine reasoning remain unexplored. 
LLMs, or more specifically, recent reasoning models such as Deepseek-R1 \citep{deepseekai2025deepseekr1incentivizingreasoningcapability} and Qwen3 \citep{yang2025qwen3technicalreport} generate CoT traces step-by-step, much like how human speech unfold over time. 
If we treat each reasoning step $z_i$ like a unit with surprisal $s(z_i)$, a single reasoning trace $\textbf{z}$ = $[z_1$, $z_2$, \dots, $z_N]$ can be analyzed in the same way to have the total effort:
\begin{equation*}
    \mathrm{Reasoning Effort}(\textbf{z}) \propto \sum_{n=1}^{N} s(z_n).    
\end{equation*}
Here, a natural question arises: \textit{does UID hypothesis hold for good reasoning patterns in LLMs?} 
A smooth, uniform surprisal profile may reflect clear and logical reasoning, while sharp spikes may signal confusion or errors. 
We extend UID hypothesis beyond psycholinguistics to probe the structure of CoT reasoning of LLMs, offering a new lens on why reasoning models succeed or fail.

\subsection{Preliminary analysis with per-step information density scores of reasoning traces}

We start by defining the step-level information density $ID_i$ for a reasoning trace $\textbf{z}=[z_1,\dots,z_N]$ with $N$ steps, 
where each reasoning step $z_i$ is composed of $M_i$ tokens, \textit{i.e.}, $z_i=[x_1,\dots,x_{M_i}]$.
We divide the reasoning steps of a single trace of a reasoning model, Qwen3-8B, by \texttt{\textbackslash n\textbackslash n}, following \citet{lightman2023letsverifystepstep}. 
Then, let $p_t(v)$ be the predictive distribution over the vocabulary at the token position $t$, and $l_t = \log p_t(x_t)$ the log-probability of the generated token $x_t$. 
To characterize $ID_i$, we consider three metrics over tokens in each step, as defined below.
\subsubsection{Three metrics of $ID_i$}
In this work, we consider three metrics for $ID_i$: (1) \textit{log-probability} $LP_i$ as a confidence signal, composed from the average token log-probability over step $i$, (2) \textit{entropy} $H_i$ as an uncertainty signal, and (3) \textit{confidence gap} $D_i$ as divergence signal defined as the difference between the log-probability of the current and the previous step. 
Details of the metrics are given in Appendix~\ref{app:idmetrics}.

\begin{figure}[h!]
    \centering
    \begin{subfigure}[t]{0.48\linewidth}
        \centering
        \includegraphics[width=\linewidth]{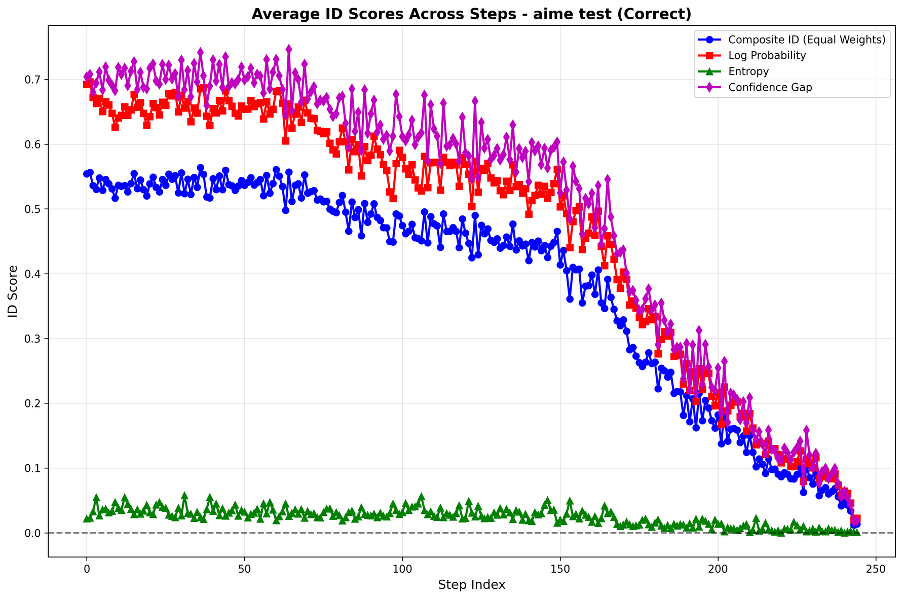}
        \caption{Correct Traces}
        \label{fig:correct}
    \end{subfigure}
    \hfill
    \begin{subfigure}[t]{0.48\linewidth}
        \centering
        \includegraphics[width=\linewidth]{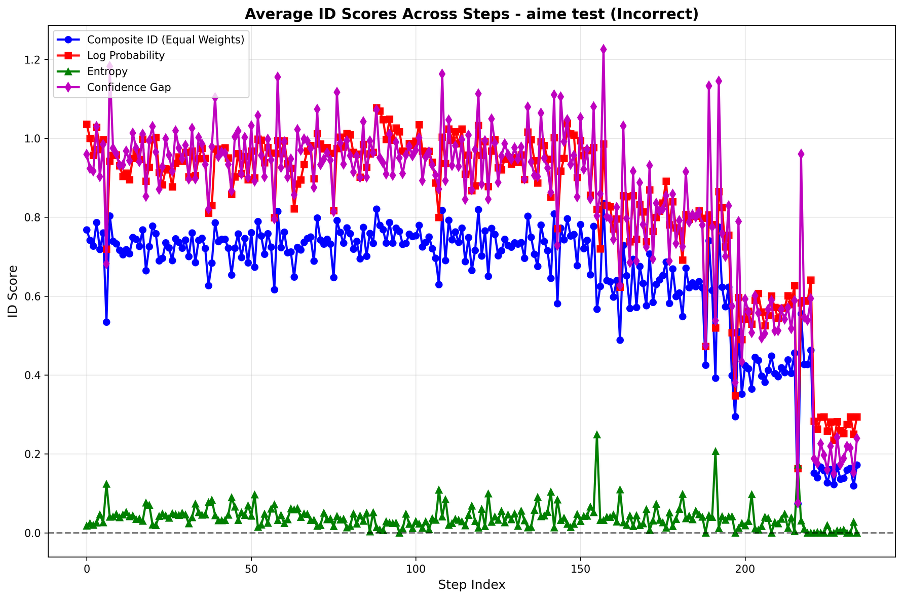}
        \caption{Incorrect Traces}
        \label{fig:incorrect}
    \end{subfigure}

    \caption{Averaged $ID$ scores ranging of correct and incorrect traces on AIME2025 test set tracked with step-level information density.}
    \label{fig:side-by-side}
\end{figure}

\subsubsection{Interpretation of the three metrics of $ID_i$ across a reasoning trace}
Figure~\ref{fig:side-by-side} compares the evolution of the three metrics -- log-probability $LP_i$, entropy $H_i$, and confidence gap $D_i$ -- and its composite metric, across reasoning traces for correct and incorrect solutions on AIME2025. 
For correct traces (Figure~\ref{fig:correct}), $H_i$ remains consistently low, while $LP_i$ and $D_i$ steadily decrease, forming a smooth trajectory that culminates in a sharp drop of the composite $ID_i$ near the final steps, to $ID_i$ score of 0.0. 
Incorrect traces (Figure~\ref{fig:incorrect}) starts higher, at average of higher than 1.0 $ID_i$ scores and show elevated and unstable $LP_i$ and $D_i$, with erratic fluctuations and sudden drops. 

\subsection{Measuring the uniformity of information density in reasoning trace}
To measure the uniformity of information density in a reasoning trace, we first clarify what "uniform" means. 
Prior psycholinguistic theory offers two perspectives \citep{meister2021revisitinguniforminformationdensity, collins2014information}. Global uniformity maintains a stable surprisal rate across the trace, while local uniformity smooth, gradual step-level changes.

Grounded in these perspectives, we explore three UID metrics for LLM reasoning traces. (1) Variance measures how much the surprisal values diverges from the mean. High variance means the reasoning process is globally unstable, with large swings in information load across steps. (2) Gini coefficient captures how unevenly the total information is distributed. A high Gini score means a few steps dominate the process, creating potential reasoning bottlenecks. (3) Shannon evenness measures how balanced the information distribution is, normalized to account for sequence length. High Shannon evenness reflects a smooth, well-balanced reasoning process. Together, these metrics distinguish between reasoning where uncertainty (entropy) is globally unstable (high variance) and unevenly concentrated (high Gini, Shannon evenness). Full definitions are in Appendix~\ref{app:uidmetrics}

%% file: 5_results.tex
\section{Unlike Human Communication, Global Non-uniform Information Distribution Predicts Reasoning Success in LLMs}
\input{Tables/MainResults}
Among the $ID_i$ metrics presented, we use entropy to compute the UID score. 
Among the three UID metrics, global uniformity, measured by variance, emerges as the strongest predictor of reasoning success. 
Selecting traces with the highest variance (low global uniformity) achieves 0.722 accuracy on AIME and 0.342 on Minerva Math, representing absolute improvements of +4.9\% and +1.6\% over the best-performing baseline (Self-Certainty: 0.689 on AIME, 0.332 on Minerva Math). 
On HMMT, high-variance traces reach 0.456 accuracy, which is +2.3\% higher than the Mean Accuracy baseline (0.433).
These results indicate that reasoning success is closely tied to low global uniformity, where models exhibit large, deliberate swings in information density throughout their thought process rather than maintaining a stable, uniform progression. Variance demonstrates consistent, cross-dataset gains, making it the most reliable signal. 
Overall, our findings suggest that reasoning is most effective when the model’s information flow is globally diverse, allowing it to shift focus dynamically and explore alternative reasoning paths, leading to stronger final answers. Experiment details are in Appendix ~\ref{app:setup}.


%% file: Tables/MainResults.tex
\begin{table*}[htbp]
  \centering
  \caption{\textbf{Main Results.} Accuracy results averaged over three random seeds.
  The best and second-best scores are bold-faced and underlined, respectively. 
  See Appendix~\ref{app:results} for more details.}
  \label{tab:MainResults}
  \scriptsize
  \resizebox{\textwidth}{!}{
  \begin{tabular}{llccc}
    \toprule
    \multirow{2}{*}{\textbf{Category}} & \multirow{2}{*}{\textbf{Method}} 
      & \textbf{AIME 2025} & \textbf{HMMT 2025} & \textbf{Minerva Math} \\
    \midrule

    \multirow{5}{*}{\textbf{Baselines}} 
      & Mean Accuracy        & 0.673 & 0.433 & 0.326 \\
      & Self-Certainty       & \underline{0.689} & \textbf{0.467} & \underline{0.332} \\
      & CoT-Decoding         & 0.678 & 0.444 & 0.330 \\
      & Highest Confidence   & 0.633 & 0.389 & 0.328 \\
      & Lowest Entropy       & 0.633 & 0.378 & 0.331 \\
    \midrule

    \multirow{2}{*}{\textbf{UID Measurement}} 
      & Highest UID Score (non-uniform) & \textbf{0.722} & \underline{0.456} & \textbf{0.342} \\
      & Lowest UID Score (uniform)      & 0.644 & 0.433 & 0.322 \\
    \bottomrule
  \end{tabular}
  }
\end{table*}

%% file: 6_conclusion.tex
\section{Conclusion}
Results show that low global uniformity strongly predicts correct reasoning, while local uniformity exhibits mixed effects. This indicates that while reasoning traces share structural similarities with natural language, their dynamics do not strictly adhere to the UID hypothesis. Instead, effective reasoning appears to rely on irregular, globally non-uniform patterns, reflecting moments of abrupt insight or decisive leaps. These findings highlight that the internal signals embedded in the structure of reasoning traces can offer valuable guidance for model design. Future work could explore how to harness these signals—rather than enforcing strict uniformity—to develop methods that adaptively leverage the natural ebb and flow of reasoning, ultimately improving the robustness and interpretability of reasoning models.

%% file: 7_appendix.tex
\appendix
\section{Implementation Details}
\subsection{Hyperparameters and GPU Setup.}
\begin{flushleft}
For all our main results, we use Qwen3-8B thinking mode. We set the temperature to 0.6, top-p to 0.95, and top-k to 20, as stated in the Qwen3 Technical Report. We use 4xA6000 GPUs for all our experiments.
\end{flushleft}

\section{Experiment Setup}\label{app:setup}
\subsection{Evaluation and Benchmarks}
\begin{flushleft}
We use accuracy for evaluation, and use three particularly challenging mathematical benchmarks, AIME2025, HMMT2025, and Minerva Math. We sample each questions five times before the final evaluation.
\subsubsection{AIME 2025.}
The American Invitational Mathematics Examination (AIME) is a prestigious US high school math contest consisting of challenging integer-answer questions. The AIME 2025 benchmark uses problems from the 2025 contests to evaluat an LLM's mathematical reasoning by requiring a single correct integer answer. The set used in our analysis contains of 30 questions.

\subsubsection{HMMT 2025.}
The Harvard-MIT Mathmematics Tournament (HMMT) is a renowned competition featuring diverse problems in algebra, geometry, combinatorics, and number theory. The HMMT 2025 benchmark uses newly released problems from the February 2025 tournament, providing a broader variety of tasks than AIME. The set used in our analysis contains of 30 questions.

\subsubsection{Minverva Math.}
The Minerva Math benchmark consists of advanced quantitative problems sourced from university-level STEM courses, including physics, chemistry, and higher mathematics. The set used in our analysis contains of 272 questions.
\end{flushleft}

\subsection{Baseline Implementation}
We re-implemeted all logic using vllm, unlike some of the codes initially released.
\subsubsection{Mean Accuracy}
This is the mean accuracy of all answers, which are sampled by 5 for each question.
\subsubsection{Self-Certainty}
This is the implementation of \citet{kang2025scalablebestofnselectionlarge}, where it first measures the confidence of each sampled answer and select one via borda-voting.
\subsubsection{CoT-Decoding}
This is the implementation of the path selection strategy used in \citet{wang2024chainofthoughtreasoningprompting}. This method, called CoT-decoding, identifies reasoning paths that contain CoT steps by measuring the model's confidence in the final answer tokens. It computes the average probability margin between the top-1 and top-2 tokens during answer decoding, denoted as $\Delta$. Decoding paths with higher $\Delta$ values are strongly correlated with correct CoT reasoning, enabling reliable extraction of CoT paths even when they are not the most probable or majority paths.

\subsubsection{Highest Confidence}
This selects the path with the highest overall token confidence in the reasoning trace.

\subsubsection{Lowest Entropy}
This selects the path with the lowest overall token entropy in the reasoning trace.

\section{Details of \texorpdfstring{$ID$}{ID} and \texorpdfstring{$UID$}{UID} Operationalizations}
\subsection{Details of \texorpdfstring{$ID_i$}{ID	extsubscript{i}} metrics} \label{app:idmetrics}
\begin{flushleft}
Log-probability $LP_i$ of a step is the average token log-probability over step $i$
\end{flushleft}
\input{Equations/Logprob}
\begin{flushleft}
Given token-level entropy $H_t$ as
\input{Equations/Tokenentropy}
\end{flushleft}
\begin{flushleft}
Step-level entropy $H_i$ is defined as   
\end{flushleft}
\input{Equations/Stepentropy}
\begin{flushleft}
Log-probability gap $D_i$ is defined as
\end{flushleft}
\input{Equations/Logprobgap}
\begin{flushleft}
Using the three metrics above, we build a composite $ID_i$ score, defined as
\input{Equations/Composite}
where all weights are equally set as $1/3$ at our current setting.
\end{flushleft}

\subsection{Averaged \texorpdfstring{\(ID\)}{ID} scores of correct and incorrect traces on HMMT2025 and Minerva Math}\label{app:steptrace}
\subsection{Mathematical Formulations of \texorpdfstring{$UID$}{UID} Operationalizations}
\begin{flushleft}\label{app:uidmetrics}
Let a reasoning trace $z$ have $N$ steps. Define the (non-negative) information density vector  
\end{flushleft}
\input{Equations/UIDvector}

\subsubsection{Operationalizing $UID(z)$ as Variance}
\begin{flushleft}

To bound $ID_i \in [0,1]$, u is normalized with min-max normalization to map the non-negative sequence to $[0,1]$.

Let
\input{Equations/Minmax}

Then, the normalized $ID_i'$ values are
\input{Equations/Minmaxnorm}

and their corresponding vector form for $UID'(z) = \tilde{\mathbf{u}} = (\mathrm{ID}'_1, \dots, \mathrm{ID}'_N)$:

Define
\input{Equations/equations}

Then, the population variance of the entries are
\input{Equations/Variance}
\end{flushleft}

\subsubsection{Operationalizing $UID(z)$ as Gini Coefficient}
\begin{flushleft}
Sort $ID_i$ values from smallest to largest, where the sorted $ID'_{1} \leq \cdots \leq ID'_{N}$
Then, the Gini coefficient can be calculated as
\end{flushleft}
\input{Equations/Gini}
\subsubsection{Operationalizing $UID(z)$ as Shannon Evenness}
\begin{flushleft}
First compute Shannon entropy of the probability normalization $p_{i} = \frac{ID_{i}}{S}$.
\input{Equations/Shannon}
with maximum $H_{\max} = \ln N$. Then, Shannon evenness can be calculated as
\input{Equations/Evenness}
\end{flushleft}
\FloatBarrier
\section{Additional Experiment Results}\label{app:results}
\subsection{Main Results with Different Seeds}
While other measures of local uniformity such as the Gini coefficient and Shannon evenness also show competitive performance, their effectieveness is limited and more dataset-dependent.
\input{Tables/Table1}
\subsection{Scaling models amplifies the role of global non-uniformity}
As shown in Table \ref{tab:scale}, scaling model size from 1.7B to 8B reveals a clear trend where variance becomes an increasingly strong predictor of reasoning success, outperforming all baselines at 8B.
\input{Tables/Table2}  

%% file: Equations/Logprob.tex
\[
LP_i = \frac{1}{b_i - a_i + 1}{\sum_{t=a_i}^{b_i} \ell_t}
\]

%% file: Equations/Tokenentropy.tex
\[
H_t = - \sum_{v \in V} p_t(v) \log p_t(v),
\]

%% file: Equations/Stepentropy.tex
\[
H_i = \frac{1}{b_i - a_i + 1}{\sum_{t=a_i}^{b_i} H_t}
\]

%% file: Equations/Logprobgap.tex
\[
D_i = {LP_i} - LP_{i-1}
\]

%% file: Equations/Composite.tex
\[
ID_i = w_{LP} {LP}_i - w_{H} {H}_i + w_{D} {D}_i
\]

%% file: Equations/UIDvector.tex
\[
\mathrm{UID}(z) = \mathbf{u} = (ID_1, ID_2, \ldots, ID_N), \quad ID_i \geq 0
\]

%% file: Equations/Minmax.tex
\[
m = \min_{1 \leq i \leq N_{\min}} \mathrm{ID}_i, \quad M = \min_{1 \leq i \leq N_{\max}} \mathrm{ID}_i
\]

%% file: Equations/Minmaxnorm.tex
\[
\mathrm{ID}'_{i} = \frac{\mathrm{ID}_{i} - m}{M - m}, \quad i = 1, \ldots, N.
\]

%% file: Equations/equations.tex
\[
S = \sum_{i=1}^{T} ID'_{i}, 
\qquad 
\mu = \frac{1}{T} \sum_{i=1}^{T} ID'_{i}, 
\qquad 
p_{i} = \frac{ID'_{i}}{S} \; \text{ (when $S > 0$)}
\]

%% file: Equations/Variance.tex
\[
\mathrm{Var}(\tilde{\mathbf{u}}) = \frac{1}{T} \sum_{i=1}^{T} \left( ID'_{i} - \mu \right)^{2}
\]

%% file: Equations/Gini.tex
\[
G(\mathbf{u}) = \frac{1}{\mu T} \sum_{i=1}^{T} (2i - T - 1)\, ID_{i}, \quad (\mu > 0).
\]

%% file: Equations/Shannon.tex
\[
H(\mathbf{u}) = - \sum_{i=1}^{T} p_{i} \ln p_{i} \quad (S > 0),
\]

%% file: Equations/Evenness.tex
\[
J'( \mathbf{u} ) = \frac{H(\mathbf{u})}{\ln T} \in [0,1].
\]

%% file: Tables/Table1.tex
\begin{table*}[htbp]
  \centering
  \caption{\textbf{Main results across various seeds}. 
  Accuracy on three mathematical benchmarks: AIME 2025, HMMT 2025, and Minerva Math.
  The \emph{Avg} sub-column reports the mean $\pm$ standard deviation across seeds.
  The best and second-best \emph{Avg} scores within each dataset block are bold-faced and underlined, respectively.}
  \label{tab:main-results}
  \resizebox{\textwidth}{!}{
  \begin{tabular}{llcccccccccccc}
    \toprule
    \multirow{2}{*}{\textbf{Category}} & \multirow{2}{*}{\textbf{Method}} 
      & \multicolumn{4}{c}{\textbf{AIME 2025}} 
      & \multicolumn{4}{c}{\textbf{HMMT 2025}} 
      & \multicolumn{4}{c}{\textbf{Minerva Math}} \\
    \cmidrule(lr){3-6} \cmidrule(lr){7-10} \cmidrule(lr){11-14}
      & & Seed 42 & Seed 1234 & Seed 2025 & Avg 
        & Seed 42 & Seed 1234 & Seed 2025 & Avg
        & Seed 42 & Seed 1234 & Seed 2025 & Avg \\
    \midrule

    \multirow{5}{*}{Baselines}
      & Mean Accuracy     & 0.680 & 0.680 & 0.660 & 0.673 $\pm$ 0.012 & 0.453 & 0.420 & 0.427 & 0.433 $\pm$ 0.017 & 0.329 & 0.325 & 0.324 & 0.326 $\pm$ 0.003 \\
      & Self-Certainty    & 0.700 & 0.633 & 0.733 & \underline{0.689 $\pm$ 0.051} & 0.433 & 0.500 & 0.467 & \underline{0.467 $\pm$ 0.034} & 0.346 & 0.331 & 0.320 & \underline{0.332 $\pm$ 0.013} \\
      & CoT-Decoding      & 0.667 & 0.667 & 0.700 & 0.678 $\pm$ 0.019 & 0.500 & 0.400 & 0.433 & 0.444 $\pm$ 0.050 & 0.335 & 0.335 & 0.320 & 0.330 $\pm$ 0.009 \\
      & Highest Confidence& 0.667 & 0.600 & 0.633 & 0.633 $\pm$ 0.034 & 0.400 & 0.367 & 0.400 & 0.389 $\pm$ 0.019 & 0.349 & 0.320 & 0.316 & 0.328 $\pm$ 0.017 \\
      & Lowest Entropy    & 0.667 & 0.600 & 0.633 & 0.633 $\pm$ 0.034 & 0.367 & 0.367 & 0.400 & 0.378 $\pm$ 0.019 & 0.349 & 0.320 & 0.324 & 0.331 $\pm$ 0.015 \\
    
    \midrule
    \multicolumn{14}{l}{\textbf{Three Measures of UID}}\\
    \addlinespace[2pt]

    Variance & Highest UID Score (non-uniform) 
             & 0.700 & 0.733 & 0.733 & \textbf{0.722 $\pm$ 0.019} & 0.467 & 0.433 & 0.467 & 0.456 $\pm$ 0.019 & 0.338 & 0.338 & 0.349 & \textbf{0.342 $\pm$ 0.006} \\
             & Lowest UID Score (uniform) 
             & 0.633 & 0.667 & 0.633 & 0.644 $\pm$ 0.019 & 0.433 & 0.433 & 0.433 & 0.433 $\pm$ 0.000 & 0.335 & 0.319 & 0.313 & 0.322 $\pm$ 0.011 \\

    Gini Coefficient & Highest UID Score (non-uniform) 
                      & 0.667 & 0.667 & 0.633 & 0.656 $\pm$ 0.019 & 0.433 & 0.333 & 0.467 & 0.411 $\pm$ 0.067 & 0.338 & 0.316 & 0.320 & 0.325 $\pm$ 0.011 \\
                      & Lowest UID Score (uniform) 
                      & 0.667 & 0.700 & 0.667 & 0.678 $\pm$ 0.019 & 0.433 & 0.433 & 0.367 & 0.411 $\pm$ 0.038 & 0.324 & 0.320 & 0.346 & 0.330 $\pm$ 0.013 \\

    Shannon Evenness  & Highest UID Score (uniform) 
                      & 0.700 & 0.667 & 0.667 & 0.678 $\pm$ 0.019 & 0.433 & 0.367 & 0.400 & 0.400 $\pm$ 0.033 & 0.320 & 0.324 & 0.331 & 0.325 $\pm$ 0.006 \\
                      & Lowest UID Score (non-uniform) 
                      & 0.633 & 0.700 & 0.600 & 0.644 $\pm$ 0.051 & 0.500 & 0.500 & 0.433 & \textbf{0.478 $\pm$ 0.039} & 0.335 & 0.320 & 0.320 & 0.325 $\pm$ 0.009 \\

    \bottomrule
  \end{tabular}}
\end{table*}

%% file: Tables/Table2.tex
\begin{table}
  \caption{\textbf{Performance across different model sizes.} 
  Performance across Qwen3-1.7B, 4B, and 8B on AIME 2025. 
  Each model is evaluated with three random seeds (42, 1234, 2025). 
  The last column within each model block shows the average across seeds. 
  The best and second-best scores are bold-faced and underlined, respectively.}
  \label{tab:scale}
  \centering
  \resizebox{\textwidth}{!}{
  \begin{tabular}{ll|cccc|cccc|cccc}
    \toprule
    \multicolumn{2}{c|}{\textbf{Baselines}} 
      & \multicolumn{4}{c|}{\textbf{1.7B}} 
      & \multicolumn{4}{c|}{\textbf{4B}} 
      & \multicolumn{4}{c}{\textbf{8B}} \\ 
    \cmidrule(lr){3-6} \cmidrule(lr){7-10} \cmidrule(lr){11-14}
    & & Seed 42 & Seed 1234 & Seed 2025 & Avg 
      & Seed 42 & Seed 1234 & Seed 2025 & Avg
      & Seed 42 & Seed 1234 & Seed 2025 & Avg \\ 
    \midrule
    & Mean Accuracy       & 0.367 & 0.353 & 0.353 & 0.358 & 0.680 & 0.653 & 0.667 & 0.667 & 0.680 & 0.680 & 0.660 & 0.673 \\
    & Self-Certainty      & 0.367 & 0.400 & 0.400 & 0.389 & 0.633 & 0.767 & 0.667 & 0.689 & 0.700 & 0.633 & 0.733 & 0.689 \\
    & CoT-Decoding        & 0.333 & 0.300 & 0.300 & 0.311 & 0.767 & 0.633 & 0.700 & 0.700 & 0.667 & 0.667 & 0.700 & 0.678 \\
    & Highest Confidence  & 0.333 & 0.367 & 0.367 & 0.356 & 0.600 & 0.567 & 0.633 & 0.600 & 0.667 & 0.600 & 0.633 & 0.633 \\
    & Lowest Entropy      & 0.333 & 0.367 & 0.367 & 0.356 & 0.567 & 0.567 & 0.633 & 0.589 & 0.667 & 0.600 & 0.633 & 0.633 \\
    \midrule
    \multicolumn{2}{c|}{\textbf{Three Measures of UID}} \\ 
    \midrule
    Variance 
      & Highest UID Score (non-uniform) & 0.433 & 0.333 & 0.333 & \textbf{0.366} & 0.667 & 0.667 & 0.700 & \textbf{0.678} & 0.700 & 0.733 & 0.733 & \textbf{0.722} \\
      & Lowest UID Score (uniform)      & 0.267 & 0.367 & 0.367 & 0.334 & 0.633 & 0.667 & 0.633 & 0.644 & 0.633 & 0.667 & 0.633 & 0.644 \\
    Gini Coefficient 
      & Highest UID Score (non-uniform) & 0.300 & 0.400 & 0.400 & \textbf{0.367} & 0.667 & 0.700 & 0.667 & 0.678 & 0.667 & 0.667 & 0.633 & 0.656 \\
      & Lowest UID Score (uniform)      & 0.367 & 0.300 & 0.300 & 0.322 & 0.700 & 0.667 & 0.700 & \textbf{0.689} & 0.667 & 0.700 & 0.667 & \textbf{0.678 }\\
    Shannon Evenness  
      & Highest UID Score (uniform)     & 0.367 & 0.267 & 0.267 & 0.300 & 0.667 & 0.567 & 0.633 & 0.622 & 0.700 & 0.667 & 0.667 & \textbf{0.678} \\
      & Lowest UID Score (non-uniform)  & 0.300 & 0.400 & 0.400 & \textbf{0.367} & 0.667 & 0.667 & 0.667 & \textbf{0.667} & 0.633 & 0.700 & 0.600 & 0.644 \\
    \bottomrule
  \end{tabular}
  }
\end{table}